\pdfoutput=1
\documentclass{article}
\usepackage[preprint, nonatbib]{neurips_2022}

\usepackage[utf8]{inputenc} 
\usepackage[T1]{fontenc}    
\usepackage{hyperref}       
\usepackage{url}            
\usepackage{booktabs}       
\usepackage{amsfonts}       
\usepackage{nicefrac}       
\usepackage{microtype}      
\usepackage{xcolor}         
\usepackage{graphicx}
\usepackage{float}

\usepackage{amsmath} 
\newtheorem{definition}{Definition}
\usepackage{stmaryrd}

\usepackage[ruled]{algorithm2e} 
\usepackage{setspace}
\SetKwComment{Comment}{/* }{ */}

\title{Learning and Compositionality: a Unification Attempt via Connectionist Probabilistic Programming}

\author{%
  Ximing Qiao\\
  Duke University\\
  \texttt{ximing.qiao@duke.edu} \\
  \And
  Hai Li\\
  Duke University\\
  \texttt{hai.li@duke.edu} \\
}

\begin{document}

\maketitle

\begin{abstract}
We consider \textit{learning} and \textit{compositionality} as the key mechanisms towards simulating human-like intelligence.
    While each mechanism is successfully achieved by neural networks and symbolic AIs, respectively, it is the combination of the two mechanisms that makes human-like intelligence possible.
    Despite the numerous attempts on building hybrid neuralsymbolic systems, we argue that our true goal should be unifying learning and compositionality, the core mechanisms, instead of neural and symbolic methods, the surface approaches to achieve them.
In this work, we review and analyze the strengths and weaknesses of neural and symbolic methods by separating their forms and meanings (structures and semantics), and propose \textit{Connectionist Probabilistic Programs} (CPPs), a framework that connects connectionist structures (for learning) and probabilistic program semantics (for compositionality).
    Under the framework, we design a CPP extension for small scale sequence modeling and provide a learning algorithm based on Bayesian inference.
Although challenges exist in learning complex patterns without supervision, our early results demonstrate CPP's successful extraction of concepts and relations from raw sequential data, an initial step towards compositional learning.
\end{abstract}

\section{Introduction}
\label{sec:intro}

Simulating human-like intelligence requires two essential mechanisms of \textit{learning} and \textit{compositionality}, while most existing algorithms only process one side well.
First, learning collects basic building blocks of knowledge from raw observational data.
    Deep learning methods based on neural networks excel at learning.
    Recent large language models absorb terabytes of text as training data, and memorize everything from language structures to commonsense knowledge~\cite{brown2020language}.
Second, compositionality allows combination of the above building blocks, which enables generalization much beyond direct observation.
    Given rules such as $A\rightarrow B$ and $B\rightarrow C$, we infer that $A\rightarrow C$ by combining them, with no actual experience on $A$ and $C$.
    This process is captured by symbolic AIs.
    Inference engines based on logic and/or probability achieve human-level question answering accuracy, given enough domain-specific knowledge bases~\cite{ferrucci2010building}.

However, it is the hybrid of the two mechanisms that makes human-like intelligence possible.
    Sticking with one mechanism alone cannot go very far.
On one side, the early failure of symbolic AIs show the obvious limitation of AIs that cannot learn.
    In logic-based expert systems, hand-writing all of commonsense as rules is obviously impossible.
    Later systems such as Bayesian networks and probabilistic programs support certain parameter learning, but still rely on human-specified core knowledge (causal relations, variable dependency, etc.)~\cite{ullman2018learning}.
On the other side, deep learning starts to hit walls, revealing the limit of learning-only systems.
    Without compositionality, neural networks only learn though direct experience, which leads to costly data collection and prolonged training.
    Models like GPT-3 still struggle to answer seemingly simple questions like ``solve for x: x + 40000 = 100000'', when the numbers are never covered during training~\cite{floridi2020gpt}.
    From a linguistics perspective, the model never truly understands the ``meaning'' of these numbers and computations.

There has been extended works on merging neural and symbolic methods (namely, \textit{neuralsymbolic} AIs~\cite{sarker2021neuro}), but this path is challenging.
At a fundamental level, the two methods express data as either distributed vectors or localized symbols, which do not fuse in any obvious ways.
As a result, successful neuralsymbolic systems generally fall to one of the two categories:
    the first is to have a neural representation as the core, and organize neural modules as symbols to simulate symbolic computation at a higher level~\cite{graves2014neural, andreas2016neural, dong2019neural};
    the second is to have a symbolic representation as the core, and wrap them by neural networks (with vector embeddings) for continuous optimization~\cite{balog2016deepcoder, mao2019neuro, shah2020learning}.
However, these methods still inherit the limitations of their neural or symbolic components.
    Symbolic organization of the neural modules cannot be efficiently learned (other than guided search), and neural-based optimization needs direct experience of all corner cases.
We still do not see learning and composition working together.

A true hybrid system should unify learning and compositionality, the core mechanisms, instead of neural and symbolic methods, which are merely surface approaches to achieve them. 
    In this work, we make an initial attempt towards this direction.
    The central idea is to breakdown neural and symbolic methods to their fundamental building blocks, then select and recombine the useful blocks.
In Section~\ref{sec:CPP}, we review and analyze the strengths and weaknesses of neural and symbolic methods by separating their forms and meanings (structures and semantics), and propose \textit{Connectionist Probabilistic Programs} (CPPs), a framework that combines a connectionist structure (for learning) with a probabilistic program semantics (for compositionality).
In Section~\ref{sec:CPPSO}, we extend the CPP focusing on a special task of sequence modeling, and provide a Bayesian inference algorithm to learn CPPs from small scale sequential data, in a compositional way.
Section~\ref{sec:result} covers initial experimental results, which reveal some challenges of learning complex patterns without supervision, but successfully demonstrate CPPs' capability to extract concepts and relations from raw data.
In Section~\ref{sec:related}~and~\ref{sec:conclusion}, we discuss related works and conclude the paper.

\section{Connectionist probabilistic programs (CPPs)}
\label{sec:CPP}

\subsection{Structure and semantics}

To motivate the design of CPPs, we first review neural and symbolic methods, and try to understand the \textit{exact components} that make them good or bad.
The method is to analyze them through separate lenses of form and meaning.
    Specifically, we denote form as \textit{structure}, referring to the data-structure stored in computer memory;
    we denote meaning as \textit{semantics}, referring to the input-output relations of functions and procedures.

For neural networks, their structures are given by weighted graphs with real-valued parameters. 
    Conventionally called connectionist networks, this type of structure is especially suited for learning due to the efficient optimization of continuous parameters.
Semantics of neural networks is given by transformations in high-dimensional vector spaces.
    While stacking a sequence of transformations gives high expressiveness, the order of stacking is fixed by the network architecture, which prohibits flexible symbolic manipulation.
The translation from the connectionist structure to the vector-space semantics is given by the McCulloch-Pitts~\cite{mcculloch1943logical} and Perceptron~\cite{gallant1990perceptron} style neuron models.
    Weights are interpreted as multiplicands to activations, and nodes are interpreted as places to perform summations and non-linear activations.
    
For symbolic AIs, the structure is given by symbols and links (pointers) that connect them, for example databases, causal graphs, and abstract syntax trees. 
    These structures are discrete, difficult to optimize, and hard to incrementally learn.
Semantics of symbolic AIs is given by mappings between symbols, capturing their relations.
    Deterministically or probabilistically, we can always build new mappings by connecting existing ones (e.g., composing two functions gives a new function).
The connection between the symbol-link structure and the relational semantics is driven by the need of human readability and programmability.
    Symbols and links are the easiest ways for human to encode their expert knowledge to a machine readable form.

The above analysis delivers two key messages. 
    First, the learning capability is given by the connectionist structure, and compositionality comes from the relational semantics.
    Second, although the vector-space semantics and symbol-link structure are helpful, they are not necessary. Neither is multiply-and-add the only model to interpret a network of weighted edges, nor symbolic expressions the only form to represent symbolic relations.
These messages motivate us to combine neural and symbolic methods at a much lower level, beneath typical neuralsymbolic approaches.

\begin{figure}
    \centering
    \includegraphics[width=\textwidth]{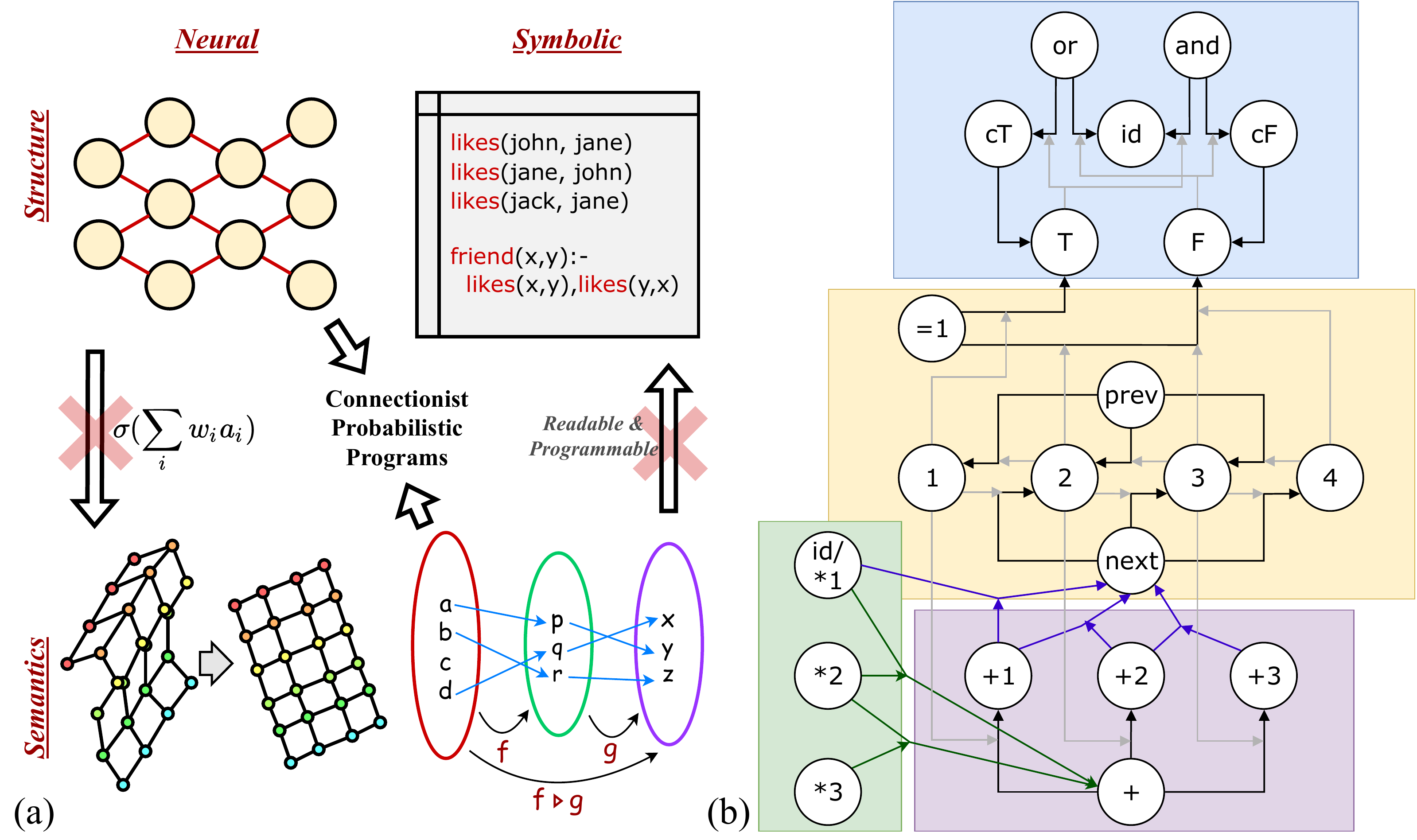}
    \caption{The structure-semantics view of CPPs and an example on logic and arithmetic.}
    \label{fig:CPP}
    \vspace{-1em}
\end{figure}

\subsection{Formal definition of CPPs}

As summarized in Figure~\ref{fig:CPP}(a), the design goal of CPPs is to drop the conventional structure-semantics connections in neural and symbolic methods, and establish a new bridge between the connectionist structure and relational semantics.
In this work, we choose the semantics of probabilistic programs over other formal systems for its close relation to human cognition~\cite{probmods2}, solid theoretical foundation~\cite{borgstrom2016lambda}, and well-studied inference algorithms~\cite{wood2014new, van2018introduction}.
Here we present the definition of CPP's structure, which includes five types of nodes/symbols and three types of weighted edges (with positivity and normalization constraints):

\begin{definition}[CPP structure]
    \label{def:CPP1}
    A CPP is a 5-tuple $\mathcal{G}=\langle Q,T,W_{Cn},W_{Fn},W_{Cm}\rangle$, where
    \begin{itemize}
        \item $Q$ is a finite set of symbols;
        \item $T:Q\rightarrow\{Id,Cn,Fn,S2,P21\}$ is a function that maps symbols to types, where $Id$ means identity functions, $Cn$ means constant functions, $Fn$ means standard functions, and $S2, P21$ are two types of combinators;
        \item $W_{Cn}:\{q\in Q:T(q)=Cn\}\rightarrow\mathbb{R}^{|Q|}_{\ge0}$ defines a probability distribution on $Q$ for every $Cn$-type symbols, and $\forall q, \|W_{Cn}(q)\|_1=1$;
        \item $W_{Fn}:\{q\in Q:T(q)=Fn\}\rightarrow\mathbb{R}^{|Q|\times|Q|}_{\ge0}$ defines a conditional probability distribution from $Q$ to $Q$ for every $Fn$-type symbols, and $\forall q, W_{Fn}(q)$ is right stochastic;
        \item $W_{Cm}:\{q\in Q:T(q)\in\{S2,P21\}\}\rightarrow\mathbb{R}^{|Q|\times|Q|}_{\ge0}$ defines a probability distribution on $Q\times Q$ for every combinator-type symbols (abbreviated as $Cm$), and $\forall q, \|W_{Cm}(q)\|_{1,1}=1$.
    \end{itemize}
\end{definition}

Unlike in neural networks, these weights define probabilistic symbolic relations (in the form of conditional probability distributions), instead of vector-space transformations.
The exact definition is given as follows (here $\delta$ refers to the Kronecker delta, and the naming of $S2$ and $P21$ originates from Figure~\ref{fig:CPPSO}(c)):
\begin{definition}[CPP declarative semantics]
    \label{def:CPP2}
    For $\mathcal{G}=\langle Q,T,W_{Cn},W_{Fn},W_{Cm}\rangle$, each symbol $q\in Q$ has a semantics $[\![q]\!]_\mathcal{G}:Q\times Q\rightarrow [0,1]$ such that:
    \begin{equation}
        \label{eqn:CPP1}
        [\![q]\!]_\mathcal{G}(i,j)=
        \begin{cases}
            \delta_{i,j}, & \text{if}\ T(q)=Id. \\
            [W_{Cn}(q)]_{j}, & \text{if}\ T(q)=Cn. \\
            [W_{Fn}(q)]_{i,j}, & \text{if}\ T(q)=Fn. \\
            \sum_{f,g}[W_{Cm}(q)]_{f,g} \llbracket f\triangleright g \rrbracket_\mathcal{G}(i,j), & \text{if}\ T(q)=S2. \\
            \sum_{f,g}[W_{Cm}(q)]_{f,g} \llbracket f\bowtie g \rrbracket_\mathcal{G}(i,j), & \text{if}\ T(q)=P21.
        \end{cases}
    \end{equation}
    where \begin{align}
        \label{eqn:CPP2}
        \llbracket f\triangleright g \rrbracket_\mathcal{G}(i,j) = & 
            \sum\nolimits_k[\![f]\!]_\mathcal{G}(i,k)[\![g]\!]_\mathcal{G}(k,j), \\
        \label{eqn:CPP3}
        \llbracket f\bowtie g \rrbracket_\mathcal{G}(i,j) = &
            \sum\nolimits_{k,l}[\![f]\!]_\mathcal{G}(i,k)[\![g]\!]_\mathcal{G}(i,l)[\![l]\!]_\mathcal{G}(k,j).
    \end{align}
\end{definition}

A symbol $q$ is understood as a probabilistic function, that randomly maps an input symbol $i$ to an output symbols $j$, with probability $[\![q]\!]_\mathcal{G}(i,j)$ (we always have $\sum_j[\![q]\!]_\mathcal{G}(i,j)\le1$, and if a function has $\sum_j[\![q]\!]_\mathcal{G}(i,j)<1$, it may not terminate).
To specify these distributions, we either directly store a probability mass function as weights in $\mathcal{G}$, as in $Cn$ and $Fn$-type symbols, or indirectly via composing other distributions, as in $S2$ and $P21$-type symbols.
    Output distributions of $Cn$-type symbols (constant functions) are not sensitive to the input $i$, thus given by $W_{Cn}(q)$.
    Output distributions of $Fn$-type symbols (standard functions) are sensitive to $i$, therefore given by $[W_{Fn}(q)]_{i,\cdot}$.
    For $S2$-type symbols, we define the composition of $f$ and $g$ (execute function $g$ after $f$) as $f\triangleright g$, and let $[\![q]\!]_\mathcal{G}(i,j)$ be a weighted mixture of all $f$'s and $g$'s, with weights $[W_{Cm}(q)]_{f,g}$.
    For $P21$-type symbols, $f\bowtie g$ defines another form of composition assuming $g$ is a higher-order function (a function that returns a function): first, execute $f$ and $g$ with a shared input $i$ to return $k$ and $l$; then execute $l$ as a function with input $k$ to return $j$.
Algorithm~\ref{alg:CPP} provides an alternative view of Definition~\ref{def:CPP2}.
    For a symbol $q$ and an input $i$, $\textsc{Sample}(q,i)$ returns $j$ with probability $[\![q]\!]_\mathcal{G}(i,j)$.

Figure~\ref{alg:CPP}(b) presents an example of CPP on expressing Boolean logic and basic arithmetic.
    Black arrows represent function outputs.
        In the blue region, $cT$ and $cF$ are constant functions (type $Cn$) that always output $T$ (logical True) and $F$ (logical False), respectively.
    Gray arrows that point to the black arrows represent function inputs.
        We have $and$ and $or$ as two standard functions (type $Fn$) that output $cT$, $cF$, or $id$ (identity function with type $Id$), depending on inputs $T$ or $F$.
        Following the above rules, a expression $and(F)(T)$ will first reduce to $cF(T)$ (since $and(F)=cF$), and then reduce to $F$ (since $cF(\cdot)=F$).
    Similarly in the yellow region, we have numbers $1\sim4$, $prev/next$ to define their order, and $=1$ to check if a number is $1$.
    Next we introduce composite functions.
        With purple arrows indicating $S2$-type composition, we have $+1=id\triangleright next,+2=+1\triangleright next$, and $+3=+2\triangleright next$.
        With green arrows denoting $P21$-type composition, we have $*1=id,*2=*1\bowtie+$, and $*3=*2\bowtie+$ (here $+$ is a higher-order function with $+(1)=+1,+(2)=+2$, and $+(3)=+3$).
        An example evaluation is $*3(1)=+(1)(*2(1))=+1(+(1)(id(1)))=+1(+1(1))=+1(2)=3$.
To simplify the illustration, Figure~\ref{alg:CPP}(b) only includes binary weights, and all functions are deterministic.
    With non-binary weights, these functions can be easily extended to probabilistic versions.

\begin{figure}[t]
\small
    \begin{minipage}{0.45\textwidth}
        \setstretch{1.039}
        \input{alg1}
    \end{minipage}
    \hfill
    \begin{minipage}{0.52\textwidth}
        \input{alg2}
    \end{minipage}
    \vspace{-1em}
\end{figure}

\section{Learning CPPs from sequential data}
\label{sec:CPPSO}

\subsection{CPPs with sequential observations (CPP-SO)}

Given the CPP as a general framework of defining complex probabilistic functions on connectionist networks, here we address the learning problem.
The idea is to extend CPPs to produce observations, such that the learn problem can be cast as Bayesian inference.
For simplicity, and to mimic the process of human obtaining knowledge from language, we focus on CPPs that generate \textit{sequential observations} (denoted as CPP-SO) on a \textit{finite, discrete alphabet}.
For an alphabet $\Gamma$ and a CPP-SO $\mathcal{G}$, the goal is to specify a probability distribution $p(x|\mathcal{G})$ for $x\in\Gamma^*$ (the Kleene closure of $\Gamma$).

Our design of CPP-SO draws inspiration from probabilistic context-free grammars (PCFGs), a classical method of sequence modeling.
We first introduce an additional type of \textit{observation ($Ob$)} symbols, mimicking the terminal nodes in PCFGs.
    These symbols behave like $Id$, but perform $\textsc{Print}$ operations (printing their \textit{labels}) as side-effects.
Next, we include eight types of combinators to handle the extra complexity of the side-effects.
As last, we add $q_0$ and $q_1$ to help defining $p(x|\mathcal{G})$:
    
\begin{definition}[CPP-SO structure]
    A CPP-SO $\mathcal{G}=\langle Q,\Gamma,T,L,q_0,q_1,W_{Cn},W_{Fn},W_{Cm}\rangle$ is a 9-tuple, where
    \begin{itemize}
        \item $Q$ is a finite set of symbols; $\Gamma$ is a finite set of alphabet letters;
        \item $T:Q\rightarrow\{Ob,Id,Cn,Fn,S1,S2,S12,S21,P1,P2,P12,P21\}$ maps symbols to types, where $Ob$ means observations, $Id,Cn,Fn$ are functions, and the rest are combinators;
        \item $L:\{q\in Q:T(q)=Ob\}\rightarrow\Gamma$ labels every $Ob$-type symbol with a letter in $\Gamma$;
        \item $q_0\in Q$ is an initial symbol; $q_1\in Q$ is an initial input;
        \item $W_{Cn}, W_{Fn}, W_{Cm}$ follow the same definition in Definition 2.
    \end{itemize}
\end{definition}

For $q\in Q$ and CPP-SO $\mathcal{G}$, we denote the semantics of $q$ as $\llparenthesis q\rrparenthesis_\mathcal{G}:Q\times\Gamma^*\times Q\rightarrow [0,1]$.
    Following Algorithm~\ref{alg:CPPSO}, we define $\llparenthesis q\rrparenthesis_\mathcal{G}(i,x,j)$ as the probability of $\textsc{Sample\&Print}(q,i)$ to print $x$ and return $j$.
To define the observation distribution, we fix the initial symbol $q_0$ and input $q_1$, then ignore the output.
    For an observation sequence $x\in\Gamma^*$, its probability is given by $p(x|\mathcal{G})=\sum_j\llparenthesis q_0\rrparenthesis_\mathcal{G}(q_1,x,j)$.

\begin{figure}
    \centering
    \includegraphics[width=\textwidth]{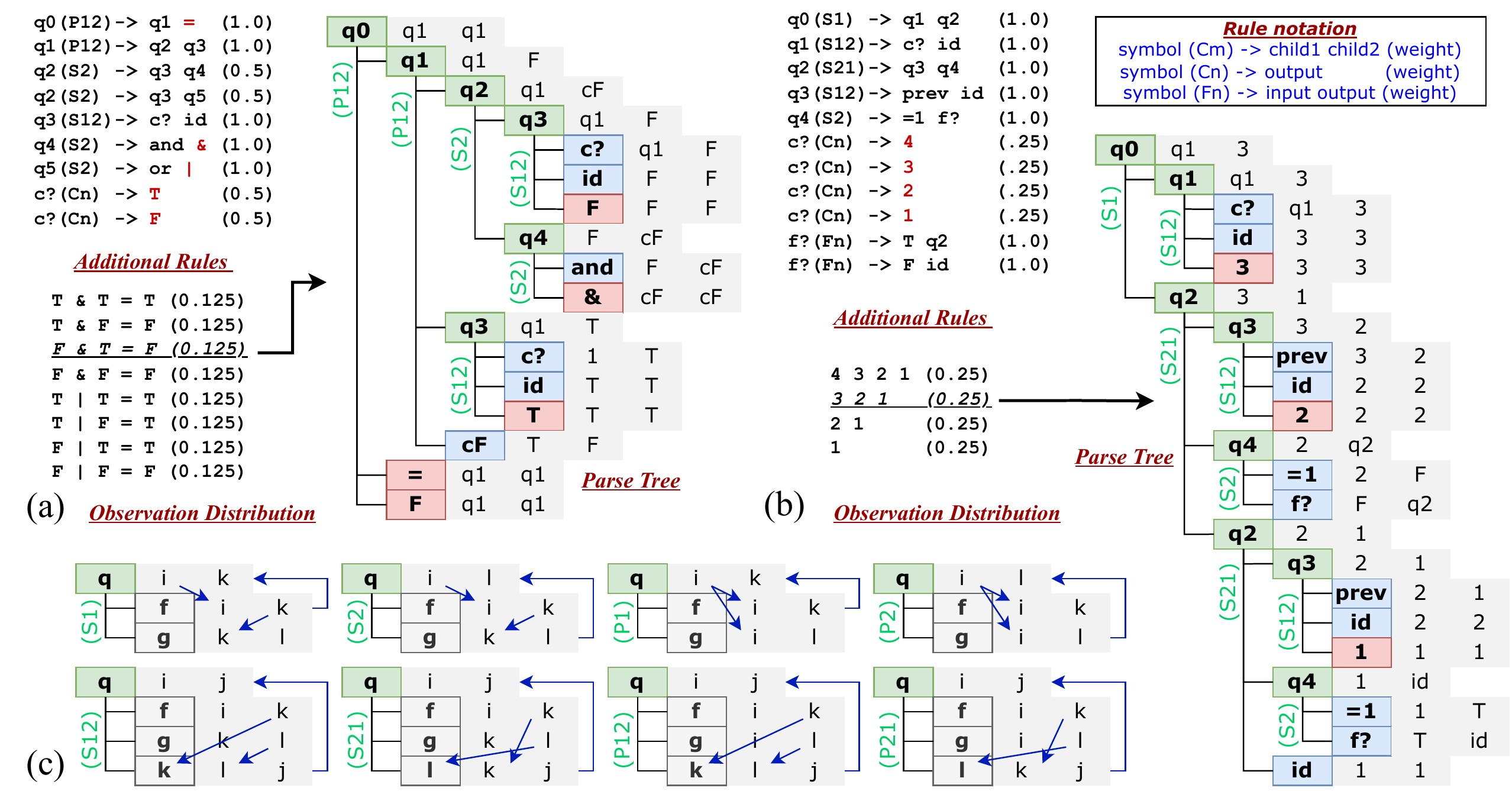}
    \caption{CPP-SO examples of (a) Boolean logic, (b) counting, and (c) basic combinators.}
    \label{fig:CPPSO}
    \vspace{-1em}
\end{figure}

Figure~\ref{fig:CPPSO}(a, b) gives two CPP-SO examples, each including a list of additional rules added to Figure~\ref{fig:CPP}(b), a observation distribution defined by these rules, and an execution trace of Algorithm~\ref{alg:CPPSO}.
    The execution traces are plotted as trees.
        They are analogous to parse trees in PCFGs, but have extra annotations of symbol types (green nodes and texts for $Cm$, blue nodes for $Cn,Fn,Id$, and red nodes for $Ob$), inputs (the first values after tree nodes), and outputs (the second values).
        Inputs and outputs of different tree nodes are connected by combinators, as illustrated by the arrows in Figure~\ref{fig:CPPSO}(c).
        Different combinators give different information flows.
    Each recursive call of $\textsc{Sample\&Print}$ advances the tree depth by one, and the $\textsc{Print}$ results (from red nodes) are read from top to bottom.

We see that CPP-SO can produce much richer observation distributions than standard PCFGs.
    The extra information flow of inputs and outputs allows a parse tree to carry complex computations, many of them hidden from observations.
In the logic example, $q3$ reads an $F$ and passes it to $q4$, which calls $and$ to produce a $cF$; then, another $q3$ reads a $T$, passes it to the $cF$ to finish the computation (recall that $and(F)(T)=cF(T)=F$), and lets $q0$ to print the final result.
In the counting example, $q1$ first generates a random number in $1\sim4$, prints it, and passes it to $q2$; $q2$ subtracts the number by 1, prints it, compares it with 1, and calls itself to start a recursion unless the number reaches 1.

\subsection{Learning as Bayesian inference (work in progress)}
\label{subsec:learning}

Given the model $p(x|\mathcal{G})$ defined above, learning becomes a Bayesian inference problem of estimating $p(\mathcal{G}|\mathbf{x})\propto p(\mathbf{x}|\mathcal{G})p(\mathcal{G})= \prod_{i=1}^n p(x_i|\mathcal{G})p(\mathcal{G})$, where $\mathbf{x}$ is a dataset of $n$ i.i.d.\ observation sequences, and $p(\mathcal{G})$ specifies the prior of $\mathcal{G}$'s parameters. 
    We assume that $Q,\Gamma,T,L,q_0,q_1$ are fixed, and $W_{Cn}, W_{Fn}, W_{Cm}$ follow Dirichlet distribution priors with a common concentration parameter $\alpha$. 
Our inference algorithm is based on a collapsed Gibbs sampler originally designed for PCFG learning~\cite{johnson2007bayesian}, but we replace its Hastings sampler with Particle Gibbs~\cite{chopin2015particle} to avoid the impractical dynamic programming parsing
(it can be derived that the complexity grows from $O(|x|^3|Q|^2)$ for PCFGs to an unacceptable $O(|x|^4|Q|^6)$ for CPP-SO, where $|x|$ is the sequence length and $|Q|$ is the number of symbols).

The first step is to introduce parse trees as auxiliary variables and collapse the parameters of $\mathcal{G}$.
    For a dataset $\mathbf{x}$, we introduce $\mathbf{t}$, a set of parse trees, where $t_i$ yields observation sequence $x_i$.
Thanks to the conjugate prior, $p(\mathbf{t}|\mathbf{x},\alpha)$ provides a sufficient statistics to $p(\mathcal{G}|\mathbf{x})$, such that we only need to sample $p(\mathbf{t}|\mathbf{x},\alpha)$ without considering $\mathcal{G}$.
    We have $p(W_{Cn}(q)|\mathbf{x})\sim Dirichlet(\boldsymbol{\alpha}+N_\mathbf{t}^{Cn}(q,\cdot))$ where $N_\mathbf{t}^{Cn}(q,j)$ is the number of times that a symbol $q$ produces an output $j$ in $\mathbf{t}$, and $\boldsymbol{\alpha}$ is a vector of $\alpha$'s.
    Likewise, $p([W_{Fn}(q)]_{i,\cdot}|\mathbf{x})\sim Dirichlet(\boldsymbol{\alpha}+N_\mathbf{t}^{Fn}(q,i,\cdot))$ where $N_\mathbf{t}^{Fn}(q,i,j)$ is the number of a symbol $q$ having an input $i$ and an output $j$.
    At last, $p(W_{Cm}(q)|\mathbf{x})\sim Dirichlet(\boldsymbol{\alpha}+N_\mathbf{t}^{Cm}(q,\cdot,\cdot))$ where $N_\mathbf{t}^{Cm}(q,f,g)$ is the number of a symbol $q$ having a pair of children $f$ and $g$.

Sampling from $p(\mathbf{t}|\mathbf{x},\alpha)$ is achieved by Gibbs sampling, where we iteratively sample $p(t_i|x_i, \mathbf{t}_{-i},\alpha)$, with $\mathbf{t}_{-i}=(t_1,\dots,t_{i-1},t_{t+1},\dots,t_n)$.
Directly computing $p(t_i|x_i, \mathbf{t}_{-i},\alpha)$ is difficult, but we can approximate it with Particle Gibbs.
    When data $x_i$ is newly encountered with no known parse $t_i$, we instantiate a Particle Filter with $M$ particles.
        This includes $M$ independent instances of Algorithm~\ref{alg:CPPSO} running in parallel, and we execute them until all perform a $\textsc{Print}$ operation.
        Then, we compare the printed labels with $x_i$, kill the particles that produce wrong observations, and duplicate the remaining particles to fill the pool of $M$ particles.
        This process is repeated until all particles terminate, and all observations are matched.
        The remaining particles give an approximate distribution of $p(t_i|x_i, \mathbf{t}_{-i},\alpha)$, which we can draw samples from.
    Once collecting an initial sample of $t_i$, later updates on $t_i$ will be driven by Conditional Particle Filters, where the $t_i$ before an update will be kept in the particle pool for conditioning. 
        This overcomes the approximation error of standard Particle Filters and allows the Gibbs sampler to mix to the target distribution $p(\mathbf{t}|\mathbf{x},\alpha)$ for any $M>1$.

Note that the presented algorithm is designed for simplicity and practicality first, and not optimized for efficiency.
    Left for future works, replacing the top-down parser (Algorithm~\ref{alg:CPPSO}) with a bottom-up parser and adding ancestor sampling~\cite{meent2015particle} to Particle Gibbs should both significantly improve the performance.

\section{Results and discussion}
\label{sec:result}

Here we present the experimental results of some toy examples and discuss the outcomes.

\begin{figure}
    \centering
    \includegraphics[width=\textwidth]{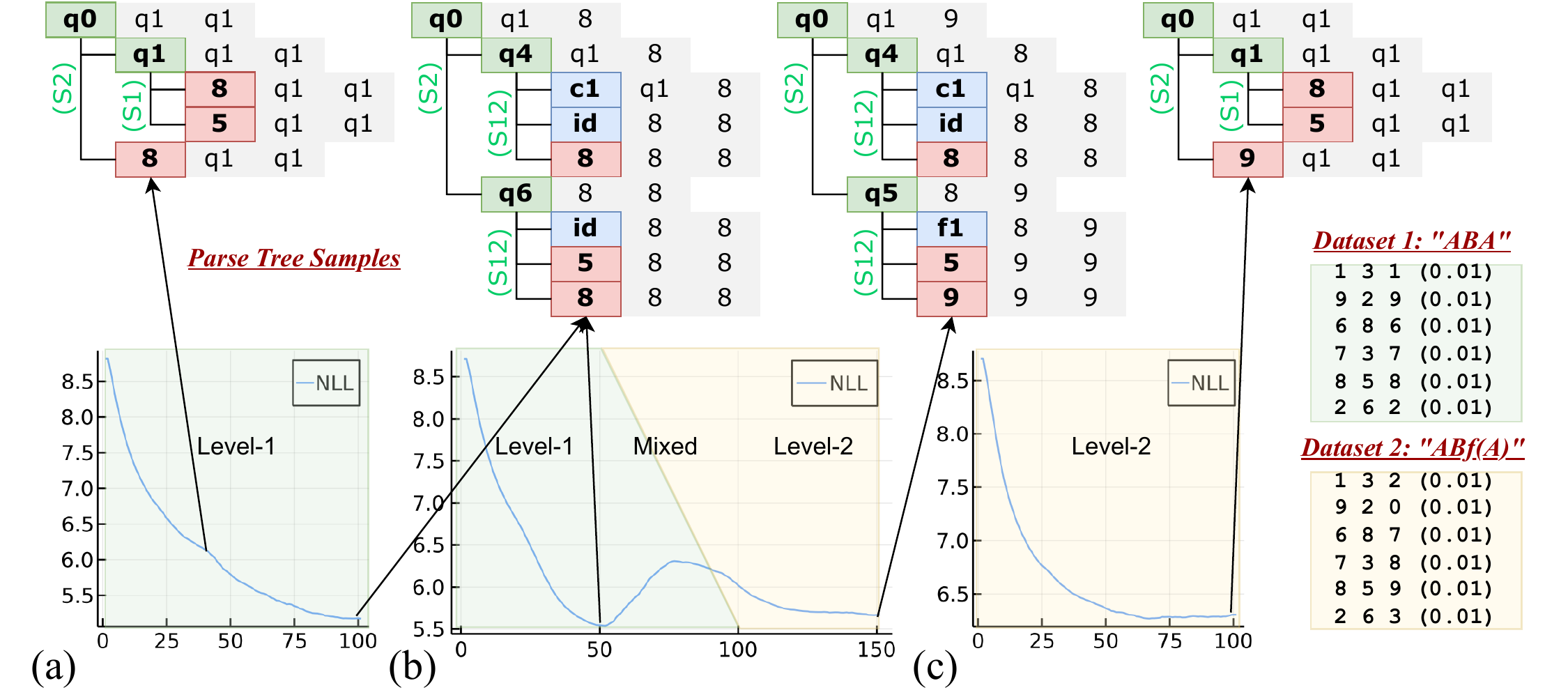}
    \caption{Dataset samples, parse tree samples in different training phases, and training curves of (a) concept learning, (b) relation learning with a curriculum, (c) relation learning without a curriculum.}
    \label{fig:result}
    \vspace{-1em}
\end{figure}

\textbf{Concept learning.}
The first step of learning is to abstract \textit{concepts} from raw signals.
    In CPP-SO, concepts are formulated as outputs of $Cn$-type nodes, with can be used for further computations (as the $T,F,3,2,1$ in Figure~\ref{fig:CPPSO}).
However, learning such an abstraction is not straightforward, since when an observation can be generated directly (as the $=,\&,|$ in Figure~\ref{fig:CPPSO}(a)), there is no reason to use an extra $Cn$ node and a larger tree.
To force the abstraction, we design a dataset that includes sentences with \textit{repeating elements}.
    Let A, B be random words, the sentences all follow the form of ``ABA''.
    This is motivated by~\cite{romberg2010statistical}, where the authors use the same technique to teach human infants to learn abstract patterns.
Without loss of generality, we use single-character words (digits in $0\sim9$) to reduce the parsing difficulty for faster learning.

From Figure~\ref{fig:result}(a), we see that in 100 epochs of Gibbs sampling, the model's training exhibits two phases, each having a sharp drop in predictive negative log-likelihood (NLL).
    In phase 1, the model learns a context-free grammar to capture the sentences, without realizing the connection between the first and the last words.
    In phase 2, the model learns a more complex rule with a $Cn$-type node $c1$, correctly identifying the ``ABA'' pattern.

\textbf{Relation learning.}
The next step of learning is to discover relations among the known concepts.
    In CPP-SO, relations are captures by input-output distributions of $Fn$-type nodes.
The challenge of relation learning is that it happens only \textit{after} concept learning.
    To achieve that, we resort to the technique of Curriculum Learning~\cite{bengio2009curriculum}, where we separate a dataset in to multiple levels, and feed data progressively from lower levels to higher levels.

Here our goal is to learn a function $f(x)=(x+1)\%10$ ($\%$ stands for modulo).
Figure~\ref{fig:result}(b) shows a model trained with 50 epochs of level-1 data (of pattern ``ABA''), 50 epochs of mixed data (level-1 data progressively replaced by level-2 data), and 50 epochs of level-2 data (of pattern ``ABf(A)'').
    After some jitters in the initial data mixing period, we see that the model correctly captures the $(x+1)\%10$ relation in a $Fn$-type node $f1$.
However, as seen from Figure~\ref{fig:result}(c), if we start directly with level-2 data without a level-1 warm up, the model will fail to capture the correct relation (note the higher NLL in (c) than in (b)).

\textbf{Knowledge composition.}
At last, we consider how to compose existing relations to new relations, and generalize to unseen cases.
    Here our target is to learn $g(x)=(x+2)\%10=(f\triangleright f)(x)$.
    A naive attempt would be to introduce a level-3 data of pattern ``ABg(A)'', however, the learned model will not generalize: it directly memorize $(x+2)\%10$ using a $Fn$ node, without using a $Cm$ node to capture the composition.
    Then, if A is restricted to digits $0\sim6$ during training, the prediction will fail when letting $A=7\sim9$ during testing.
See Appendix for more discussions on challenges and solutions to obtain the correct composition.

\textbf{Discussion.}
The experiments reveal several limitations of our current learning method.
One issue is that since a model must master one level of curriculum before advancing to the next level, if the curriculum is long, the model is unlikely to pass all levels.
    This can be systematically solved by having a ``meta'' Particle Filter over the Gibbs sampler: to simulate multiple MCMC chains in parallel, and resample after each curriculum level.
A deeper issue is about the curriculum design, as can be observed from the complex datasets used in the above experiments.
    A good curriculum must supply the right form of data at the right level, and update smoothly to guide a model's progress.
    Without a good curriculum, the model either fails to learn, or learn the wrong pattern.

We believe the curriculum design is an intrinsic challenge of learning complex patterns without supervision.
Human learners circumvent this issue by sensing a wide range of non-language clues.
    Human infants learn numbers by seeing visual objects, addition by counting fingers, multiplication by teachers' instructions.
    By comparison, our text-only training is equivalent to a human trying to learn math with exercise questions only: no teachers, no discussions, even no textbooks.
Perhaps a more natural approach of learning is to begin with control instead of language problems, like how human infants learn to move before to speak. 
    Control problems have ``built-in'' curriculum automatically generated by the progressive exploration of more complex environments and actions.

Despite the limitations, our results show that CPP-SO includes all necessary components for compositional learning, and works as a small scale functioning prototype.
A bigger lesson from CPP-SO it that we can jump out of the long-standing dichotomy of neural and symbolic methods.
    Backpropagation through matrix multiplications is not the only way to train connectionist networks: probabilistic methods can be equally useful (as in Section~\ref{subsec:learning}).
    Symbolic relations may not be limited in a human readable/programmable form: a continuous relaxation can still maintain a well-defined semantics (as in Definition~\ref{def:CPP2}).
By freely recombining structures and semantics of existing systems, we have a rich world of new models to explore.

\section{Related works}
\label{sec:related}
\vspace{-.2em}

\textbf{Neuralsymbolic methods.}
Besides the two categories introduced in Section~\ref{sec:intro}, some neuralsymbolic methods do exhibit connectionist structures and relational semantics.
Tensor symbolic representation~\cite{smolensky1990tensor} use tensor products to capture relations, at the cost of expensive computation (similar to Equation~\ref{eqn:CPP3}).
    We circumvent the dense summation by particle-based sparse sampling and make the algorithm computationally practical. 
Logical neural networks~\cite{riegel2020logical} use a first-order logic semantics, which leads to different inference algorithms.

\textbf{Probabilistic program synthesis.}
Our work is closely related to probabilistic program synthesis~\cite{saad2019bayesian, ellis2020dreamcoder, lake2015human}, due to the similar semantics and inference algorithms.
However, existing works focus on solving problems in restricted domains with domain-specific languages.
Our method is designed for general sequence modeling, which aims to potentially model general cognition and human-like language-driven learning.

\textbf{Probabilistic neuron models.}
Since the Bayesian brain hypothesis, there has been numerous models of probabilistic interpretations of neural coding, although the applications have been limited to simple perceptual and control problems~\cite{doya2007bayesian, ma2014neural}.
Our connectionist formulation of probabilistic programs might provide possible ways to bridge probabilistic neuron models to high level reasoning.

\textbf{Formal semantics theory.}
The design of CPP-SO is heavily inspired by the semantics theory of natural language~\cite{coppock2019invitation}, in particular~\cite{steedman2011combinatory}, where meaning of (sub-)sentences is given by combinators and higher-order functions.
We extend the lexicalized grammar (where all functions are attached to words) to support hidden computations (with non-observed sub-trees) for more flexible modeling.


\section{Conclusion}
\label{sec:conclusion}
\vspace{-.2em}

We present CPP and CPP-SO as methods to express and learn compositional knowledge.
Our initial experiments show their capability to extract concepts and relations from raw sequential data.
Although learning complex patterns is limited by the difficult curriculum design, the functioning of our method suggests the potential of rethinking and recombining the structural and semantic components of classical algorithms.
As for future works, we are looking for efficient inference algorithm (based on bottom-up parsing) that scales well to natural language processing. 
Meanwhile, by extending the grammar to model observation-action sequences, we will apply it to reinforcement learning and optimal control problems, which avoids the curriculum design issue and moves closer to fully autonomous learning.
In a long term, it would be interesting to see if learning and compositionality are sufficient to simulate human-like intelligence, or there are more necessary elements to be included.

{\small
\bibliographystyle{unsrt}
\bibliography{reference}
}


\newpage
\appendix
\section{Appendix}

\subsection{Concept learning}
\label{exp:1}

In this experiment we try to learn $Cn$ nodes with ``ABA'' pattern data.

\textbf{Experiment setup:}
\begin{itemize}
    \item Dataset design: the dataset includes 100 i.i.d. samples drawn by the following algorithm: given an alphabet $\Gamma=\{0,1,2,3,4,5,6,7,8,9\}$, sample $A$ and $B$ independently and uniformly from $\Gamma$, and then return the concatenation of $string(A)$, $string(B)$, and $string(A)$ (here $string()$ converts a number, e.g. $0$, to a string ``0'').
    \item Hyperparameter setting: in addition to 10 $Ob$-type nodes (one for each alphabet letter), we include 3 nodes for $S2$ and $S12$ type each, 2 nodes for $Cn$ and $Fn$ type each, and 1 node for type $Id$. Initial symbol $q_0$ and input $q_1$ are both set to a $S2$-type node. The weights all follow Dirichlet priors with $\alpha=0.1$, except for one special element in $W_{Cm}$, as described next. The number of particles is 100.
    \item Special initialization: denote a $S12$-type node $q_4$, a $Cn$-type node $c_1$, the $Id$-type node $id$, we raise the $\alpha$ for $[W_{Cm}(q_4)]_{c_1,id}$ to 100. This special initialization forces a high initial probability for $q_4$ to expand to $c_1$ and $id$.
    \item Training procedure: we train the model on the described dataset for 100 epochs.
    \item Performance metric: we report the negative log likelihood (NLL) of the observation sequence normalized by the sequence length. Note the NLL cannot be computed directly, and we use the Particle Filter weights to get an unbiased estimation. In the plots, we apply an exponential smoothing to the normalized NLL with rate $0.9$.
\end{itemize}

\textbf{Results:}
We show the NLL-epoch plots for 10 independent MCMC runs. 
Each run takes around 15 seconds on a 4-core CPU.
Despite the different shape of learning curves, all models successfully captures the ``ABA'' pattern.
\begin{figure}[H]
    \centering
    \includegraphics[width=0.8\textwidth]{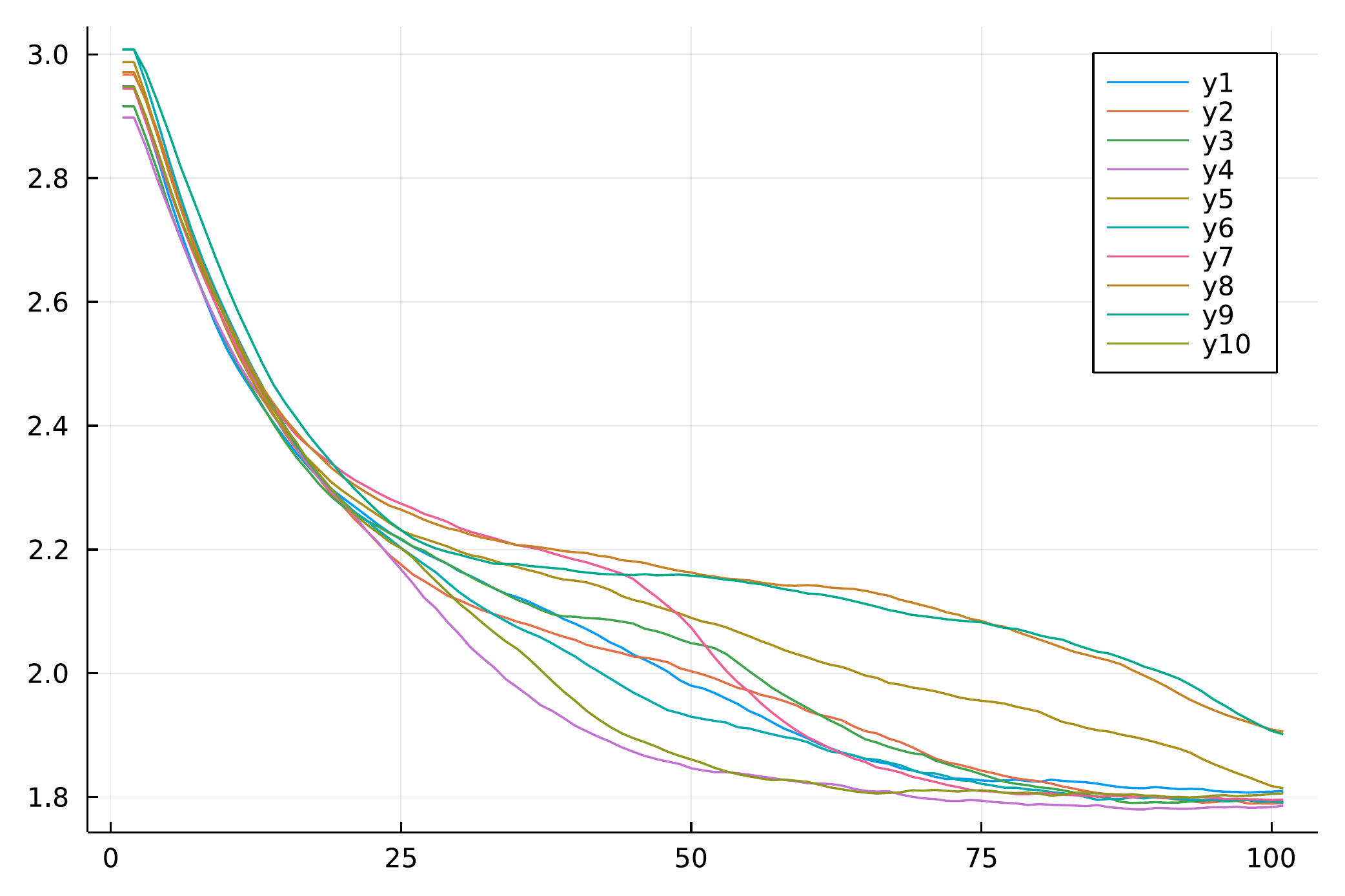}
\end{figure}

\newpage
\subsection{Relation learning}
\label{exp:2}

In this experiment we try to learn $Fn$ nodes with ``ABf(A)'' pattern data.

\textbf{Experiment setup:}
\begin{itemize}
    \item Dataset design: the level-1 dataset is the same as above. The level-2 dataset includes 100 i.i.d. samples drawn by the following algorithm: given an alphabet $\Gamma=\{0,1,2,3,4,5,6,7,8,9\}$, sample $A$ and $B$ independently and uniformly from $\Gamma$, then return the concatenation of $string(A)$, $string(B)$, and $string(f(A))$ (here $string()$ converts a number, e.g. $0$, to a string ``0'', and $f(x)=(x+1)\%10$).
    \item Hyperparameter setting: the same prior distribution as above. The number of particles is expanded to 400.
    \item Special initialization: the same as above.
    \item Training procedure: we first train the model on the level-1 dataset for 50 epochs. Then among the next 50 epochs of training, in each epoch we randomly select 2 level-1 data entries and replace them with 2 level-2 data entries. At last, we train the model on the fully replaced dataset (only including level-3 data) for another 50 epochs.
    \item Performance metric: the same as above.
\end{itemize}

\textbf{Results:}
We show the NLL-epoch plots for 10 independent MCMC runs. 
Each run takes around 80 seconds on a 4-core CPU. 
Whether the model successfully learns $f$ depends on the exact way it captures the level-1 data. 
    The learning will succeed if level-1 data are captured by a tree structure that include an $Id$ node outside of $q_4$ (e.g., the one in Figure~\ref{fig:result}(a), epoch 100), since this $Id$ node can be replaced by an $Fn$ node during level-2 learning. 
    For tree structures that do not include an $Id$ node outside of $q_4$, the learning will fail.
\begin{figure}[H]
    \centering
    \includegraphics[width=0.8\textwidth]{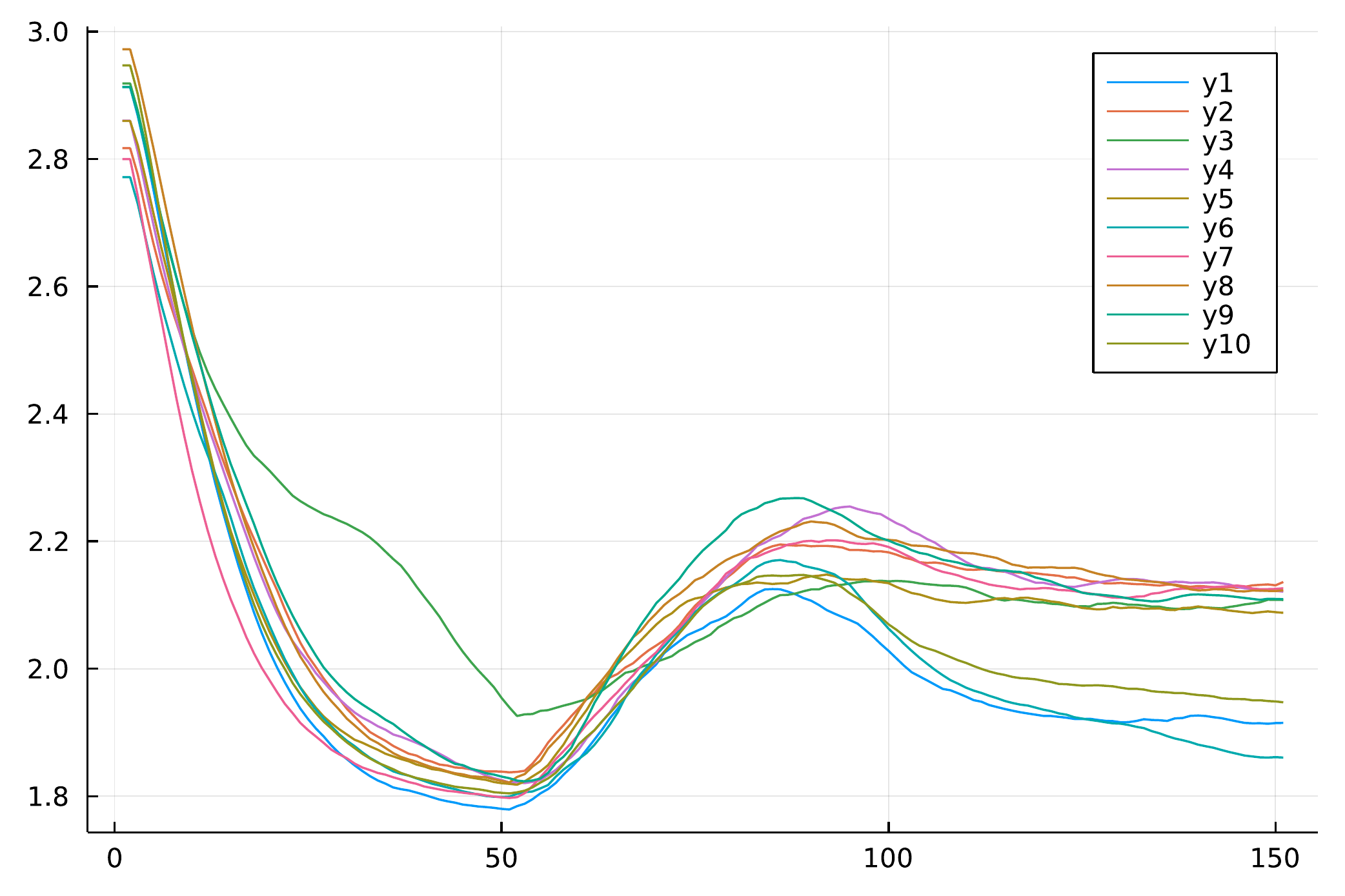}
\end{figure}

\newpage
\subsection{Relation learning (alternative method)}
\label{exp:3}

In this experiment we provide a more efficient (higher success rate) way to learn $Fn$ nodes with ``ABAf(A)'' pattern data.

\textbf{Experiment setup:}
\begin{itemize}
    \item Dataset design: the level-1 dataset is the same as in Experiment~\ref{exp:1}. The level-2 dataset includes 100 i.i.d. samples drawn by the following algorithm: given an alphabet $\Gamma=\{0,1,2,3,4,5,6,7,8,9\}$, sample $A$ and $B$ independently and uniformly from $\Gamma$, then return the concatenation of $string(A)$, $string(B)$, $string(A)$, and $string(f(A))$ (here $string()$ converts a number, e.g. $0$, to a string ``0'', and $f(x)=(x+1)\%10$).
    \item Hyperparameter setting: the same as above.
    \item Special initialization: the same as above.
    \item Training procedure: the same as above.
    \item Performance metric: the same as above.
\end{itemize}

\textbf{Results:}
We show the NLL-epoch plots for 10 independent MCMC runs. 
Each run takes around 100 seconds on a 4-core CPU. 
By use a longer observation sequence to better guide the learning, we find an increased success rate of learning the right relation (note that due to the longer sequence, the normalized NLLs can be lower than those in Experiment~\ref{exp:2}).
\begin{figure}[H]
    \centering
    \includegraphics[width=0.8\textwidth]{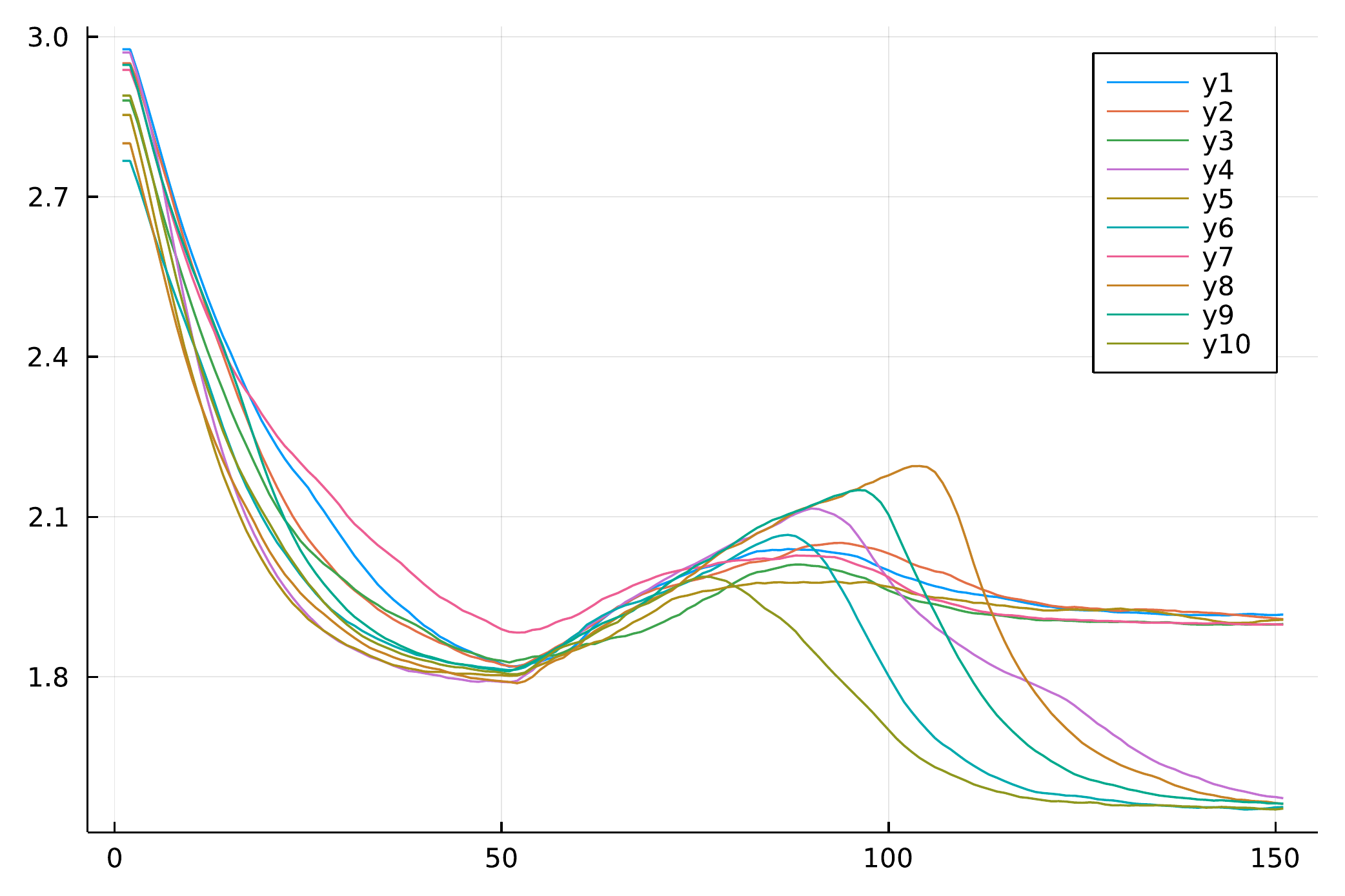}
\end{figure}

\newpage
\subsection{Knowledge composition (naive method, failed)}
\label{exp:4}

In this experiment we try to learn a relation $f$ with ``ABf(A)'' data first, then learn $g$ with ``ABg(A)'' data where $g=f\triangleright f$. However, the model directly memorizes the $g$ as a $Fn$ node without learning the composition $f\triangleright f$.

\textbf{Experiment setup:}
\begin{itemize}
    \item Dataset design: the level-1 and level-2 datasets are the same as in Experiment~\ref{exp:1} and~\ref{exp:2}, respectively. The level-3 dataset includes 100 i.i.d. samples drawn by the following algorithm: given an alphabet $\Gamma=\{0,1,2,3,4,5,6,7,8,9\}$, sample $A$ and $B$ independently and uniformly from $\Gamma$, then return the concatenation of $string(A)$, $string(B)$, and $string(g(A))$ (here $string()$ converts a number, e.g. $0$, to a string ``0'', and $g(x)=(x+2)\%10$).
    \item Hyperparameter setting: the same as above.
    \item Special initialization: the same as above.
    \item Training procedure: we first train the model on level-1 data for 50 epochs, then mixed level-1 and 2 data for 50 epochs (following the same mixing policy as in Experiment~\ref{exp:2}), then level-2 data for 50 epochs, then mixed level-2 and 3 data for 50 epochs, at last level-3 data for 50 epochs.
    \item Performance metric: the same as above.
\end{itemize}

\textbf{Results:}
We show the NLL-epoch plots for 10 independent MCMC runs. 
Each run takes around 150 seconds on a 4-core CPU.
We see that if a model successfully learns level-2 data, it will also succeeds level-3.
However, by analyzing the learned model's weights, we find that once level-2 data are replace by level-3 data, the model forgets the relation $f$, and replaces it by the relation $g$. 
Therefore, without a good dataset design, the model fails to learn the right composition.
\begin{figure}[H]
    \centering
    \includegraphics[width=0.8\textwidth]{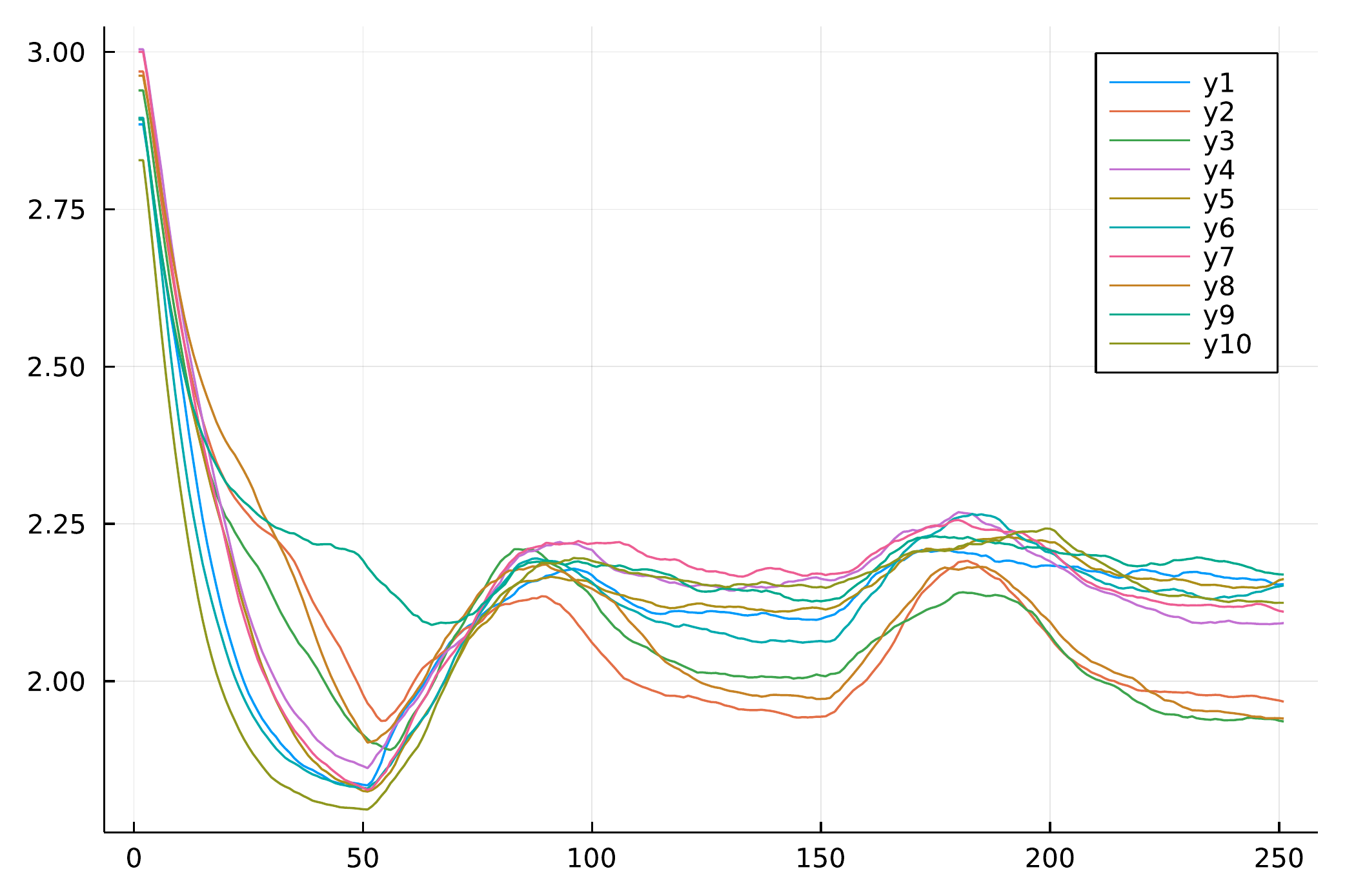}
\end{figure}

\newpage
\subsection{Knowledge composition (alternative method 1, failed)}
\label{exp:5}

In this experiment we try to provide a better guide to the learning with ``ABAf(A)'' data first, and then ``ABAf(A)g(f(A))'' data later. However, the model still cannot learn the composition.

\textbf{Experiment setup:}
\begin{itemize}
    \item Dataset design: the level-1 and level-2 datasets are the same as in Experiment~\ref{exp:1} and~\ref{exp:3}, respectively. The level-3 dataset includes 100 i.i.d. samples drawn by the following algorithm: given an alphabet $\Gamma=\{0,1,2,3,4,5,6,7,8,9\}$, sample $A$ and $B$ independently and uniformly from $\Gamma$, then return the concatenation of $string(A)$, $string(B)$, $string(A)$, $string(f(A))$, and $string(g(f(A)))$ (here $string()$ converts a number, e.g. $0$, to a string ``0'', and $g(x)=(x+2)\%10$).
    \item Hyperparameter setting: the same as above.
    \item Special initialization: the same as above.
    \item Training procedure: the same as above.
    \item Performance metric: the same as above.
\end{itemize}

\textbf{Results:}
We show the NLL-epoch plots for 10 independent MCMC runs. 
Each run takes around 200 seconds on a 4-core CPU.
The idea of this dataset design is to force the model learn $g$ without forgetting $f$, since it must first apply $f$ to produce $f(A)$, and then $g$ to produce $g(f(A))$. 
    In theory, the model should capture the composition $g=f\triangleright f$ if the MCMC chain mixes. 
    However, we do not observe that in practice.
    By analyzing the learned weights, we see the model always learns two separate $Fn$ nodes, where one captures $f$ and the other captures $g$.
\begin{figure}[H]
    \centering
    \includegraphics[width=0.8\textwidth]{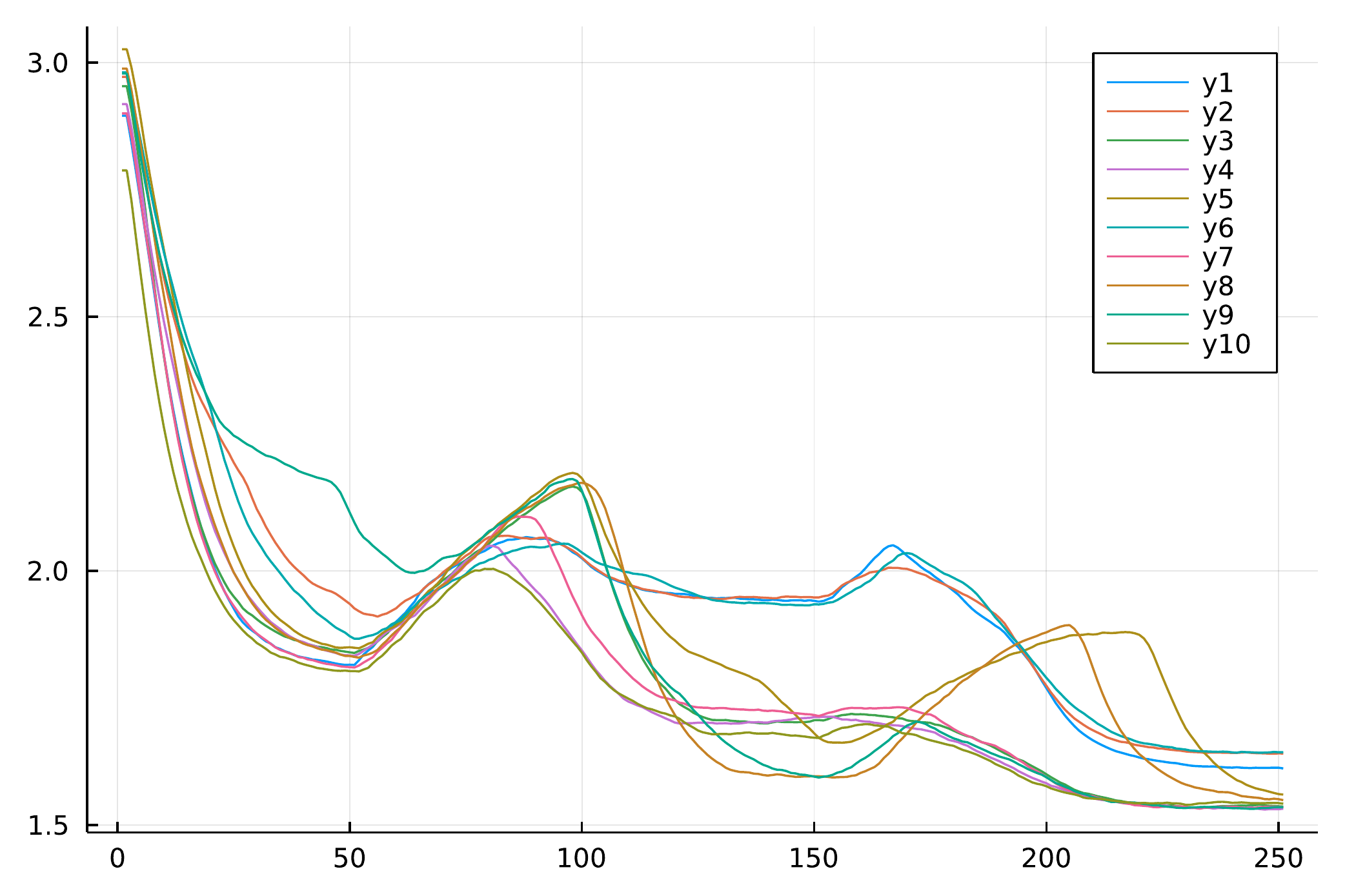}
\end{figure}

\newpage
\subsection{Knowledge composition (alternative method 2, failed)}
\label{exp:6}

In this experiment we directly encode $f$ to the prior. We find that the failure of composition learning is due to the inefficient top-down parser from Algorithm~2, instead of the dataset or curriculum design.

\textbf{Experiment setup:}
\begin{itemize}
    \item Dataset design: the same as in Experiment~\ref{exp:4}.
    \item Hyperparameter setting: the same as above.
    \item Special initialization: in addition to the one in Experiment~\ref{exp:1}, we select a $Fn$-type node $f_1$, and set the prior of $W_{Fn}(f_1)$ to capture the relation $f(x)=(x+1)\%10$, i.e., the $\alpha$ of $[W_{Fn}(f_1)]_{i,j}$ is set to 100 if $L(j)=f(L(i))$, and 0.1 otherwise.
    \item Training procedure: similar to Experiment~\ref{exp:2}, but using level-1 and level-3 datasets, instead of level-1 and level-2.
    \item Performance metric: the same as above.
\end{itemize}

\textbf{Results:}
We show the NLL-epoch plots for 10 independent MCMC runs. 
Each run takes around 100 seconds on a 4-core CPU.
Here we simplify Experiment~\ref{exp:5} by directly initialize the relation $f$ as a prior, without learning it from a curriculum.
    However, learning the composition still fails in all 10 runs.
    
Our analysis shows that the top-down parser of Algorithm~2 is extremely inefficient in identifying novel function compositions: it must guess the first $f$ correctly by pure chance, then the second $f$ correctly by pure chance again, until composing them to $g$. 
    This implies that the probability of making the correctly guess scales inverse-squared to the number of symbols.
    Thus, although learning the correct composition should happen if we continue the MCMC forever, we never see it happens in the first 100 epochs of Gibbs sampling.
\begin{figure}[H]
    \centering
    \includegraphics[width=0.8\textwidth]{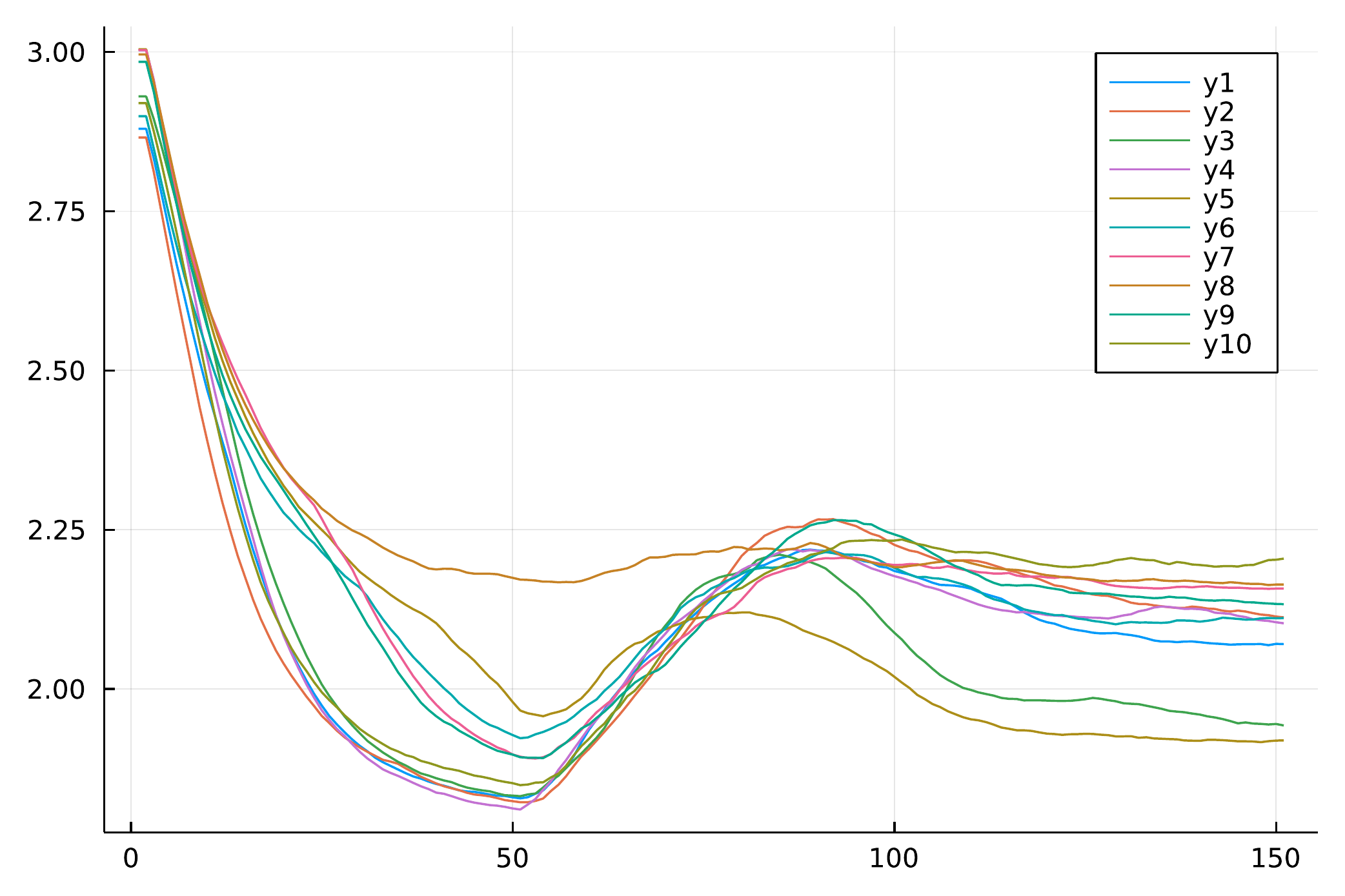}
\end{figure}

\newpage
\subsection{Knowledge composition (alternative method 3, succeeded)}
\label{exp:7}

As a temporary fix, here we adjust the prior to successful guide the composition learning. 
Meanwhile, our initial analysis (not presented here) shows that the working-in-progress bottom-up parser can systematically solve this problem without relying to such ad-hoc prior adjustments. The difficulty of composition learning found in the above experiments is not a fundamental problem of the CPP-SO framework, and will be solved soon in future works.

\textbf{Experiment setup:}
\begin{itemize}
    \item Dataset design: the same as above.
    \item Hyperparameter setting: the same as above.
    \item Special initialization: in addition to the one in Experiment~\ref{exp:6}, we select a $S2$-type node $q_2$, and set the prior of $[W_{Cm}(q_2)]_{f_1,f_1}$ to 100, forcing it to capture the composition $f\triangleright f$.
    \item Training procedure: the same as above.
    \item Performance metric: the same as above.
\end{itemize}

\textbf{Results:}
We show the NLL-epoch plots for 10 independent MCMC runs. 
Each run takes around 100 seconds on a 4-core CPU.
In this experiment, we adjust the prior to force the symbol $q_2$ to have a high probability of composing two $f_1$'s.
    The resulting model correctly learns the desired composition whenever successfully modeling the data.
\begin{figure}[H]
    \centering
    \includegraphics[width=0.8\textwidth]{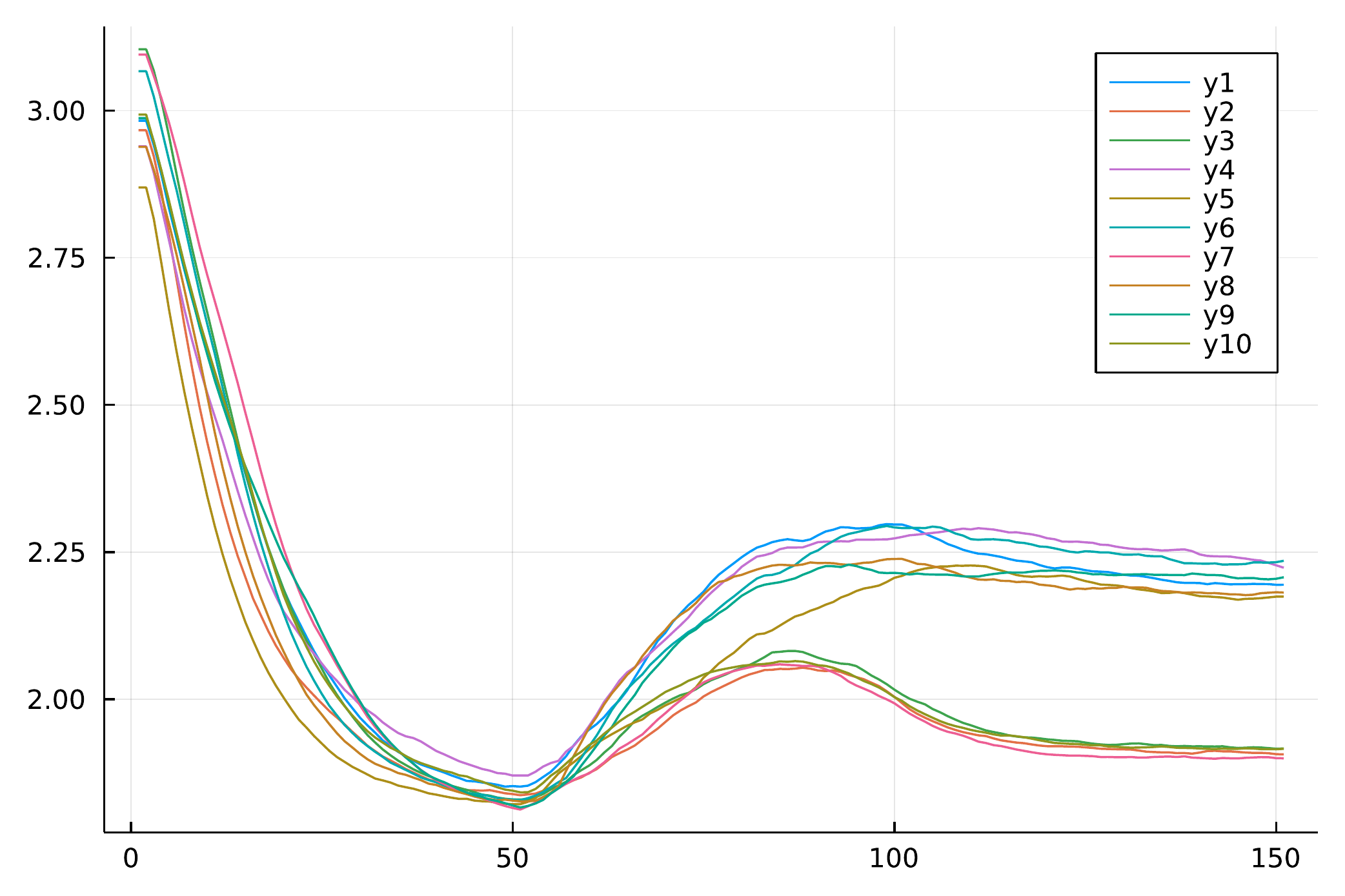}
\end{figure}

\end{document}